\documentclass[10pt,twocolumn,letterpaper]{article}

\usepackage{iccv}
\usepackage{times}
\usepackage{epsfig}
\usepackage{graphicx}
\usepackage{amsmath}
\usepackage{amssymb}
\usepackage{algorithm}
\usepackage[utf8]{inputenc}
\usepackage{soul}
\usepackage{booktabs}
\usepackage{subcaption}
\usepackage{array}
\usepackage[font=small,skip=0pt]{caption}

\usepackage{algpseudocode}
\DeclareMathOperator*{\argmin}{arg\,min}

\usepackage[pagebackref=true,breaklinks=true,letterpaper=true,colorlinks,bookmarks=false]{hyperref}

\iccvfinalcopy % *** Uncomment this line for the final submission

\def\iccvPaperID{1353} % *** Enter the ICCV Paper ID here

\newcolumntype{A}{>{\centering\arraybackslash}p{5.5em}}
\newcolumntype{C}{>{\centering\arraybackslash}p{2em}}
\begin{document}

%%%%%%%%% TITLE
\title{Learning from Video and Text via Large-Scale Discriminative Clustering}

\author{Antoine Miech$^{1,2}$ \quad Jean-Baptiste Alayrac$^{1,2}$ \quad Piotr Bojanowski$^{2}$ \quad Ivan Laptev\,$^{1,2}$ \quad Josef Sivic$^{1,2,3}$\\
$^1$École Normale Supérieure\thanks{$^1$Département d’informatique de l’ENS, École normale supérieure, CNRS, PSL Research University, 75005 Paris, France.} \quad \quad \quad $^2$Inria\quad \quad \quad $^3$CIIRC\thanks{$^3$Czech Institute of Informatics, Robotics and Cybernetics at the Czech Technical University in Prague.}\\
}
\renewcommand\footnotemark{}

\maketitle
%\thispagestyle{empty}
%\iffalse
\begin{abstract}

  Discriminative clustering has been successfully applied to a number of weakly-supervised learning tasks. 
  Such applications include person and action recognition, text-to-video alignment, object co-segmentation and co-localization in videos and images. 
  One drawback of discriminative clustering, however, is its limited scalability.
  We address this issue and propose an online optimization algorithm based on the Block-Coordinate Frank-Wolfe algorithm.
  We apply the proposed method to the problem of weakly-supervised learning of actions and actors from movies together with corresponding movie scripts.
  The scaling up of the learning problem to 66 feature-length movies enables us to significantly improve weakly-supervised action recognition.
 \end{abstract}

%%%%%%%%%%%%%%%%%%%%
%%% INTRODUCTION %%%
%%%%%%%%%%%%%%%%%%%%

\section{Introduction}

Action recognition has been significantly improved in recent years. Most existing methods~\cite{laptev08learning,simonyan2014,wang13action,wang2015}
rely on supervised learning and, therefore, require large-scale, diverse and representative action datasets for training.
Collecting such datasets, however, is a difficult task given the high costs of manual search and annotation of the video.
Notably, the largest action datasets today are still orders of magnitude smaller (UCF101~\cite{soomro2012}, ActivityNet~\cite{caba2015activitynet})
compared to large image datasets, they often contain label noise and target specific domains such as sports (Sports1M~\cite{karpathy2014}).
%Charades  \cite{sigurdsson2016}

%Automatic action and person recognition is a very active field of research that has seen important advances in the last years.
%However, training robust action models require important amounts of annotated data.
%Annotating videos with human actions is a very time consuming process, as it requires to identify the action, and provide it's spatial and temporal extent.
%As opposed to objects, it is hard to properly define the spatial and temporal extent of an action, which makes the manual annotation a tedious task.

Weakly supervised learning aims to bypass the need of manually-annotated datasets using readily-available, but possibly noisy and incomplete supervision.
Examples of such methods include learning of person names from image captions or video scripts~\cite{berg2004names,everingham06hello,sivic09who,tapaswi12knock}.
Learning actions from movies and movie scripts has been approached in~\cite{bojanowski13finding,bojanowski14weakly,duchenne09automatic,laptev08learning}.
Most of the work on weakly supervised person and action learning, however, has been limited to one or a few movies. Therefore the power of leveraging large-scale weakly annotated
video data has not been fully explored.

\begin{figure}[t]
  \mbox{}\vspace{.3cm}\\
  \begin{center}
  \includegraphics[width=\linewidth]{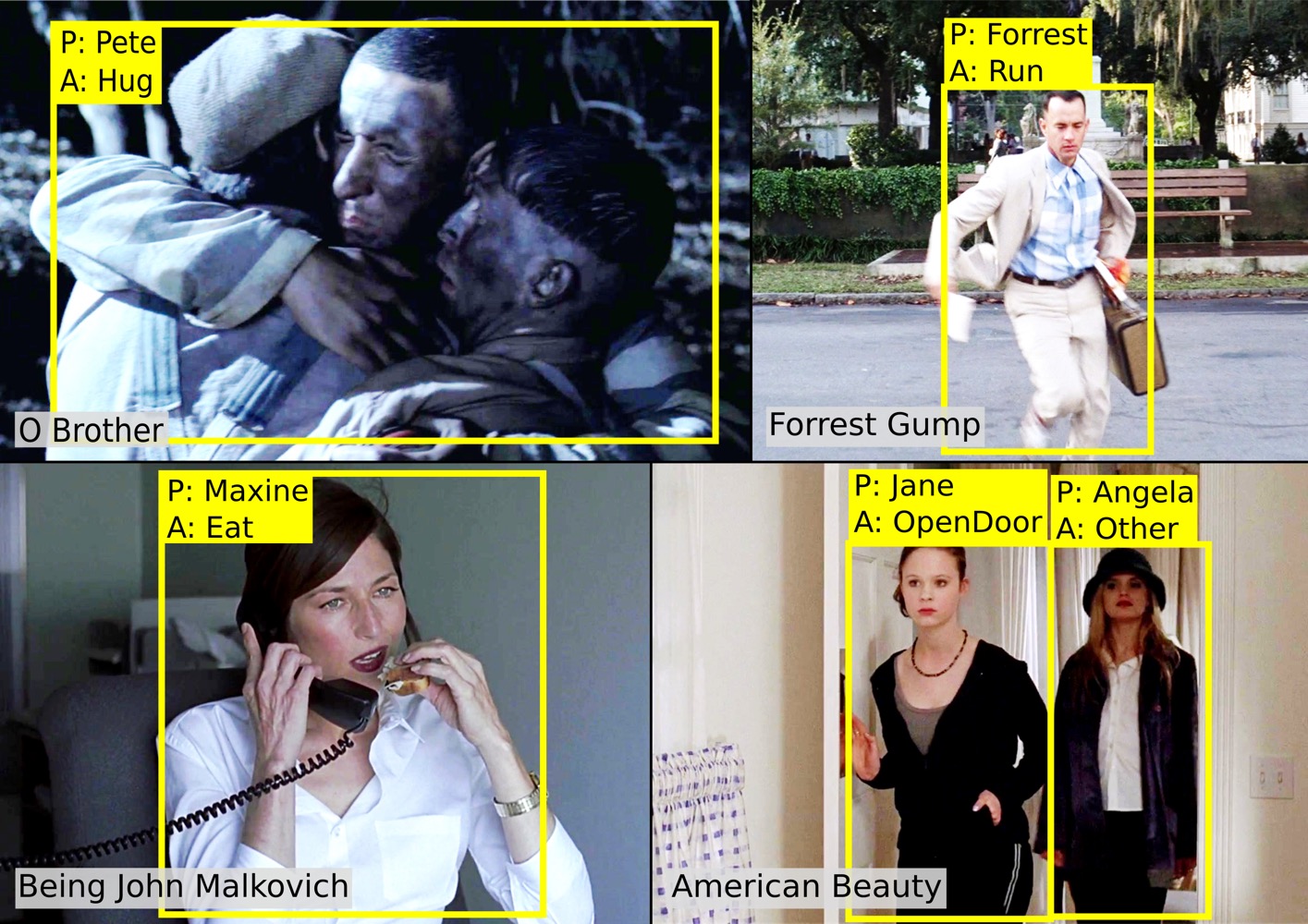}
\end{center}
\caption{We automatically recognize actors and their actions in a of dataset of 66 movies with scripts as weak supervision.}
\label{fig:teaser}
\vspace{-.5cm}
\end{figure}

In this work we aim to scale weakly supervised learning of actions.
In particular, we follow the work of~\cite{bojanowski13finding} and learn actor names together with their actions from movies and movie scripts.
While actors are learned separately for each movie, differently from~\cite{bojanowski13finding}, our method simultaneously learns actions from all movies and movie scripts available for training.
Such an approach, however, requires solving a large-scale optimization problem.
We address this issue and propose to scale weakly supervised learning by adapting the Block-Coordinate Frank-Wolfe approach~\cite{lacoste13bcfw}.
Our optimization procedure enables action learning from tens of movies and thousands of action samples, readily available from our subset of movies or other recent datasets with movie descriptions~\cite{rohrbach15dataset}.
This, in turn, results in large improvements in action recognition.

Besides the optimization, our work introduces a new model for background class in the form of a constraint. It enables better and automatic modeling of the background class (\ie unknown actors and actions).
We evaluate our method on 66 movies and demonstrate  significant improvements for both actor and action recognition.
Example results are illustrated in Figure~\ref{fig:teaser}.

\subsection{Related Work}
This section reviews related work on discriminative clustering, Frank-Wolfe optimization and its applications to the weakly supervised learning of people and actions in video.
%
%weakly-supervised person and action recognition in video together with the work on Frank-Wolf optimization.
%We review in this section recent work that is most closely related to our model.

\vspace{-.4cm}
\paragraph{Discriminative clustering and Frank-Wolfe.}
%In order to solve our optimization problem, which is a very large quadratic problem, we use the Frank-Wolfe algorithm~\cite{frank56algorithm,jaggi13revisiting}.
The Frank-Wolfe algorithm~\cite{frank56algorithm,jaggi13revisiting} allows to minimize large convex problems over convex sets by solving a sequence of linear problems.
In computer vision, it has been used in combination with discriminative clustering~\cite{bach07diffrac} for action localization~\cite{bojanowski14weakly}, text-to-video alignment~\cite{alayrac16unsupervised,bojanowski15weakly}, object co-localization in videos and images~\cite{tang14efficient} or instance-level segmentation~\cite{seguin16instance}. % dependency-tree induction~\cite{grave15convex} 
A variant of Frank-Wolfe with randomized block coordinate descent was proposed in~\cite{lacoste13bcfw}.
This extension leads to lower complexity in terms of time and memory requirements while preserving the convergence rate.
In this work we build on~\cite{lacoste13bcfw} and adapt it for the problem of large-scale weakly supervised learning of actions from movies.
%Recent extensions of~\cite{lacoste13bcfw} proposed in~\cite{osokin16bcfw} improve convergence speed have been proposed in~\cite{osokin16bcfw}.
%Among them, sampling strategies based on the decomposable duality gap improved speed of convergence in the case of the structured SVM problem.
%Frank-Wolfe algorithm was successfully applied in computer vision and natural language processing. % especially in the context of weak supervision.
%Methods optimizing  discriminative clustering cost functions with linear models and quadratic losses lead to quadratic optimization problems.
%The Frank-Wolfe algorithm is particularly well suited for this kind of models and it has been used in the context of action localization~\cite{bojanowski14weakly}, text-to-video alignment~\cite{bojanowski15weakly}, co-localization in videos and images~\cite{tang14efficient}, dependency-tree induction~\cite{grave15convex} or instance-level segmentation~\cite{seguin16instance}.

\vspace{-.4cm}
\paragraph{Weakly supervised action recognition.}
Movie scripts are used as a source of weak supervision for temporal action localization in~\cite{duchenne09automatic}.
An extension of this work~\cite{bojanowski14weakly} exploits the temporal order of actions as a learning constraint.
Other~\cite{lan11discriminative} target spatio-temporal action localization and recognition in video using a latent SVM.
A weakly supervised extension of this method~\cite{shapovalova12similarity} localizes actions without location supervision at the training time.
Another recent work~\cite{weinzaepfel16towards} proposes a multi-fold Multiple-Instance Learning (MIL) SVM to localize actions given video-level supervision at training time.
Closer to us is the work of~\cite{bojanowski13finding} that improves weakly supervised action recognition by joint action-actor constraints derived from scripts.
While the approach in~\cite{bojanowski13finding} is limited to a few action classes and movies, we propose here a scalable solution and demonstrate significant improvements in action recognition when applied to the large-scale weakly supervised learning of actions from many movies.

\vspace{-.4cm}
\paragraph{Weakly supervised person recognition.}
Person recognition in TV series has been studied in~\cite{everingham06hello,sivic09who} where the authors propose a solution to the problem of associating speaker names in scripts and faces in videos. Speakers in videos are identified by detecting face tracks with lip motion.
The method in~\cite{cour09learning} presents an alternative solution by formulating the association problem using a convex surrogate loss. 
Parkhi \etal~\cite{parkhi15it} present a method for person recognition combining a MIL SVM with a model for the background class.
Most similar to our model is the one presented in~\cite{bojanowski13finding}.
The authors propose a discriminative clustering cost under linear constraints derived from scripts to recover the identities and actions of people in movies.
Apart from scaling-up the approach of~\cite{bojanowski13finding} to much larger datasets, our model extends and improves~\cite{bojanowski13finding} with a new background constraint.
\vspace{-.3cm}

%\begin{figure*}[t]
%\begin{center}
%\includegraphics[width=0.9\textwidth,trim={0 1.9cm 1.3cm 2.1cm},clip]{overview}
%\end{center}
%\caption{Overview} 
%\end{figure*}

%\begin{figure*}[t]
%\begin{center}
%\includegraphics[width=0.9\textwidth,trim={0 1.9cm 1.3cm 2.1cm},clip]{overview}
%\end{center}
%\caption{Overview} 
%\end{figure*}

\paragraph{Contributions.}
In this paper we make the following contributions:
(i) We propose an optimization algorithm based on Block-Coordinate Frank-Wolfe that allows scaling up discriminative clustering models~\cite{bach07diffrac} to much larger datasets.
(ii) We extend the joint weakly-supervised Person-Action model of~\cite{bojanowski13finding}, with a simple yet efficient model of the background class. 
(iii) We apply the proposed optimization algorithm to scale-up discriminative clustering to an order of magnitude larger dataset, resulting in significantly improved action recognition performance.

%%%%%%%%%%%%%
%%% MODEL %%%
%%%%%%%%%%%%%

%\section{Discriminative Clustering for Weakly-Supervised model}
\section{Discriminative Clustering for Weak Supervision}

We want to assign labels (\eg names or action classes) to samples (\eg person tracks in the video).
As opposed to the standard supervised learning setup, the exact labels of samples are not known at training time.
Instead, we are given only partial information that some samples in a subset (or bag) may belong to some of the labels.
This ambiguous setup, also known as multiple instance learning, is common, for example, when learning human actions from videos and associated text descriptions.

To address this challenge of ambiguous and partial labels, we build on the discriminative clustering criterion based on a linear classifier and a quadratic loss (DIFFRAC~\cite{bach07diffrac}).
This framework has shown promising results in weakly supervised and unsupervised computer vision tasks~\cite{alayrac16unsupervised,bojanowski13finding,bojanowski14weakly,bojanowski15weakly,joulin10discriminative,joulin12multi,ramanathan14linking,seguin16instance}. 
In particular, we use this approach to group samples into linearly separable clusters. % separable by a linear model. 
Suppose we have $N$ samples to group into $K$ classes.
We are given $d$-dimensional features $X \in \mathbb{R}^{N\times d}$, one for each of the $N$ samples, and our goal is to find a binary matrix 
$Y \in \{0,1\}^{N\times K}$ assigning each of the $N$ samples to one of the labels, 
where  $Y_{nk} = 1$ if and only if the sample $n$ (\eg~a person track in a movie) is assigned to the label $k$ (\eg~action class running). 

%\begin{align}
%  \label{eq:constraint-row}
%Z1_{P} = 1_{N},
%\end{align}
%where $1_{N}$ (resp. $1_{P}$) is the vector of size $N$ (resp. $P$) with all coefficients at 1. We have $Z_{np} = 1$ if and only if track $n$ is assigned to person $p$. 
 \label{cost}
First, suppose the assignment matrix $Y$ is given. In this case finding a linear classifier $W$ can be formulated as a ridge regression problem
\begin{align}
  \label{eq:W}
  \min_{W \in \mathbb{R}^{d\times K}} \frac{1}{2 N} \|Y - X W \|_{\mathrm F}^{2} + \frac{\lambda}{2} \ \| W \|_{\mathrm F}^{2},
\end{align} 
where $X$ is a matrix of input features, $Y$ is the given labels assignment matrix, $\| . \|_{\mathrm F}$ is the matrix norm 
(or Frobenius norm) induced by the matrix inner product $\langle.,.\rangle_{\mathrm F}$ (or Frobenius inner product) and $\lambda$ is a regularization hyper-parameter set to a fixed constant.  
The key observation is that we can resolve the classifier $W^*$ in closed form as
%For all $Z$, the close form solution of this minimization problem is :
\begin{equation}
\label{eq:Wmin}
  W^{*}(Y)= (X^\top X + N \lambda I)^{-1} X^\top Y.
\end{equation}
%where the solution depends only on the input features $X$, given assignment of people to tracks $Z$, number of tracks $N$ and fixed constant $\lambda$.
  
In our weakly supervised setting, however, $Y$ is unknown. 
Therefore, we treat $Y$ as a latent variable and optimize~(\ref{eq:W}) w.r.t.~$W$ and $Y$.
%This corresponds to finding the best possible assignment of people to tracks so that they become easily separable by a linear classifier $W$.   
%Our final goal is to actually retrieve the best possible matrix $Z$ we consider here  as a latent variable.
In details, plugging the optimal solution $W^*$~\eqref{eq:Wmin} in the cost~\eqref{eq:W} removes the dependency on $W$ and the cost can be written as a quadratic function of $Y$, i.e.~$C(Y) = \langle Y, A(X,\lambda) Y\rangle_{\mathrm F}$, where $A(X, \lambda)$ is a matrix that only depends on the data $X$ and a regularization parameter $\lambda$. Finding the best assignment matrix $Y$ can then be written as the minimization of the cost $C(Y)$:
\begin{align}
   \label{eq:binary-pb}
   \min_{Y \in \{0,1\}^{N \times K}} \ \langle Y, A(X,\lambda) Y\rangle_{\mathrm F}.
\end{align}

Solving the above problem, however, can lead to degenerate solutions~\cite{bach07diffrac} unless additional constraints on $Y$ are provided. 
In section~\ref{application}, we incorporate weak supervision in the form of constraints on the latent assignment matrices $Y$.
The constraints on $Y$ used for weak supervision generally decompose into small independent blocks.
This block structure is the key for our optimization approach that we will present next.

\subsection{Large-Scale optimization}
\label{large-scale-opt}

The Frank-Wolfe (FW) algorithm has been shown effective for optimizing convex relaxation of~\eqref{eq:binary-pb} in a number of vision problems~\cite{alayrac16unsupervised,bojanowski13finding,bojanowski14weakly,bojanowski15weakly,Joulin14efficient,seguin16instance}.
It only requires solving linear programs on a set of constraints.
Therefore, it avoids costly projections and allows the use of complicated constraints such as temporal ordering~\cite{bojanowski14weakly}.
However, the standard FW algorithm is not well suited to solve~\eqref{eq:binary-pb} for a large number of samples $N$.

First, storing the $N \times N$ matrix $A(X,\lambda)$ in memory becomes prohibitive (\eg the size of $A$ becomes $\geq$ 100GB for $N \geq 200000$). 
Second, each update of the FW algorithm requires a full pass over the data resulting in a space and time complexity of order $N$ for each FW step.

Weakly supervised learning is, however, largely motivated by the desire of using large-scale data with ``cheap'' and readily-available but incomplete and noisy annotation.
Scaling up weakly supervised learning to a large number of samples is, therefore, essential for its success.
We address this issue and develop an efficient version of the FW algorithm. % that scaling linearly with the number of samples.
Our solution builds on the Block-Coordinate Frank-Wolfe (BCFW)~\cite{lacoste13bcfw} algorithm and extends it with a smart block-dependent update procedure as described next.
The proposed update procedure is one of the key contribution of this paper.

\subsubsection{Block-coordinate Frank-Wolfe (BCFW)}\label{sec:BCFW}

%%When dealing with optimization problem of the form $\min_{T\in\mathcal{T}} f(T)$, where the convex feasible set $\mathcal{T}$ is compact and the function $f$ is convex, the Frank-Wolfe algorithm~\cite{frank56algorithm} is an iterative optimization procedure that only requires to optimize \emph{linear} functions over $\mathcal{T}$.
%%Each step of the algorithm consists in computing the gradient at the current iterate in order to obtain a \emph{linear} approximation of the function.
%%This approximation is then minimized on the set $\mathcal{T}$, and the solution is used to get the direction for the next update.
The Block-Coordinate version of the Frank-Wolfe algorithm~\cite{lacoste13bcfw} is useful when the convex feasible set $\mathcal{Y}$ can be written as a Cartesian product of $n$ smaller sets of constraints: $\mathcal{Y}=\mathcal{Y}^{(1)}\times\ldots\times \mathcal{Y}^{(n)}$.
Inspired by coordinate descent techniques, BCFW consists of updating one variable block $\mathcal{Y}^{(i)}$ at a time with a reduced Frank-Wolfe step.
This method has potentially $n$ times cheaper iterates both in space and time. We will see  that most
of the weakly supervised problems exhibit such a block structure on latent variables.

\subsubsection{BCFW for discriminative clustering}
\label{sec:BCFW_DIFFRAC}
To benefit from BCFW, we have to ensure that the time and space complexity of the block update does not depend on the total number of samples $N$ (\eg person tracks in all movies) but only depends on the size $N_i$ of smaller blocks of samples~$i$ (\eg person tracks within one movie).
%While it was shown to be possible for the structured SVM~\cite{lacoste13bcfw}, it does not necessarily follows for another problem.
After a block is sampled, the update consists of two steps.
First, the gradient with respect to the block is computed. 
Then the \emph{linear oracle} is called to obtain the next iterate.
As we show below, the difficult part in our case is to efficiently compute the gradient with respect to the block.

\paragraph{Block gradient: a naive approach.} \label{block-gradient}
Let's denote $N_{i}$ the size of block $i$.
The objective function $f$ of problem~\eqref{eq:binary-pb} is $f(Y)= \langle Y, A(X,\lambda) Y\rangle_{\mathrm F} $, where (see~\cite{bach07diffrac})
\begin{equation}
  A(X,\lambda) = \frac{1}{2N} (I_{N} - X(X^\top X + N \lambda I_{d})^{-1}X^\top).
\end{equation}
To avoid storing matrix $A(X,\lambda)$ of size $N \times N$, one can precompute the matrix $P=(X^\top X + N\lambda I_{d})^{-1}X^\top\in\mathbb{R}^{d\times N}$.
We can write the block gradient with respect to a subset of samples $i$ as follows:
\begin{equation}
\label{eq:block-gradient}
\nabla_{(i)}f(Y) = \frac{1}{N} (Y^{(i)} - X^{(i)}PY),
\end{equation}
where $Y^{(i)} \in \mathbb{R}^{N_{i}\times K}$ and $X^{(i)}\in \mathbb{R}^{N_{i}\times d}$ are the label assignment variable and the feature matrix for block $i$ (\eg person tracks in movie $i$), respectively.
Because of the $PY$ matrix multiplication, naively computing this formula has the complexity $\mathcal{O}(NdK)$, where $N$ is the total number of samples, $d$ is the dimensionality of the feature space and $K$ is the number of classes.
As this depends linearly on $N$, we aim to find a more efficient way to compute block gradients, as described next.

\paragraph{Block gradient: a smart update.} 
We now propose an update procedure that avoids re-computation of block gradients and whose time and space complexity at each iteration depends on $N_{i}$ instead of $N$.
%by avoiding the re-computation of the block gradients. 
The main intuition is that we need to find a way to store information about all the blocks in a compact form.
A natural way of doing so is to maintain the weights of the linear regression parameters $W \in \mathbb{R}^{d\times K}$.
From~\eqref{eq:Wmin} we have $W = PY$.
If we are able to maintain the variable $W$ at each iteration with the desired complexity $\mathcal{O}(N_idK)$, then the block gradient computation~\eqref{eq:block-gradient} can be reduced from $\mathcal{O}(NdK)$ to $\mathcal{O}(N_idK)$.
We now explain how to effectively achieve that.

At each iteration $t$ of the algorithm, we only update a block $i$ of $Y$ while keeping all other blocks fixed.
We denote the direction of the update by $D_t\in\mathbb{R}^{N\times K}$ and  the step size by $\gamma_t$.
With this notation the update becomes
\begin{equation}
\label{eq:block-update}
Y_{t+1} = Y_t + \gamma_t D_t.
\end{equation}
The update rule for the weight variable $W_t$ can now be written as follows:
\begin{equation}
\begin{split}
\label{eq:weight-update}
&W_{t+1} = P(Y_t + \gamma_t D_t) \\
&W_{t+1} = W_{t} + \gamma_{t} PD_t, 
\end{split}
\end{equation}
Recall that at iteration $t$, BCFW only updates block $i$, therefore  $D_t$ has non zero value only at block $i$. 
In block notation we can therefore write the matrix product $PD_t$ as:
\begin{equation}
\hspace*{-0.2cm}
\begin{array}{c}
   \\
   \\
\left[ P^{(1)},\cdots,P^{(i)},\cdots,P^{(n)} \right] \times \\
\end{array}
\hspace*{-0.2cm}
\left[
\begin{array}{p{0.45cm}c}
 0 \\
$D_t^{(i)}$ \\
 0 \\
\end{array}
\right]
\hspace*{-0.2cm}
\begin{array}{c}
   \\
   \\
= P^{(i)}D_t^{(i)},  \\
\end{array}
\end{equation}
where  $P^{(i)} \in \mathbb{R}^{d\times N_{i}}$ and $D_t^{(i)}\in  \mathbb{R}^{N_{i}\times K}$ are the i-th blocks of matrices 
$P$ and $D_t$, respectively. The outcome is an update of the following form
%with $P^{(i)} \in \mathbb{R}^{d\times N_{i}}$ and $D_t^{(i)}\in  \mathbb{R}^{N_{i}\times K}$.
\begin{equation}
W_{t+1} = W_{t} + \gamma_{t} P^{(i)}D_t^{(i)},
\end{equation}
where the computational complexity for updating $W$ has been reduced to $\mathcal{O}(N_{i}dK)$ compared to $\mathcal{O}(NdK)$ in
the standard update.

We have designed a Block-Coordinate Frank-Wolfe update with time and space complexity depending only on the size of the blocks and not the entire dataset.
%Thus, the BCFW method can be effectively applied to problems with large number of samples.
This allows to scale discriminative clustering to problems with a very large number of samples.
%This allows the BCFW algorithm to solve and scale to problems with a very large number of samples.
The pseudo-code for the algorithm is summarized in Algorithm~\ref{CHalgorithm}. Next, we describe an application of this large-scale discriminative clustering algorithm to weakly supervised person and action recognition in movies. 

\definecolor{comment}{RGB}{86, 115, 154}

\begin{algorithm}[t!]
\caption{BCFW for Discriminative Clustering~\cite{bach07diffrac}}
\label{CHalgorithm}
\begin{algorithmic}
\State Initiate $Y_0$, $P:=(X^\top X + N\lambda I_{d})^{-1}X^\top$, $W_0=PY_0$, $g_i=+\infty, \ \forall i$.
\For{ $t$ = 1 \ldots $N_{iter}$}
\State $i$ $\gets$ sample from distribution proportional to $\textit{g}$ \cite{osokin16bcfw}
\State $\nabla_{(i)}f(Y_t)$ $\gets$ $\frac{1}{N}(Y^{(i)}_t - X^{(i)}W_t)$ {\color{comment} \footnotesize \# Block gradient}
%\State 
\State $Y_{min}$ $\gets$ $\argmin_{x \in \mathcal{Y}^(i)} \langle \nabla_{(i)}f(Y_t), x\rangle_{\mathrm F}$     {\color{comment} \footnotesize \# Linear oracle}
%\State {\color{comment} \footnotesize \# Block gap and line-search computation}
\State $D \gets Y_{min} - Y^{(i)} $
\State $g_i \gets - \langle D, \nabla_{(i)}f(Y_t)\rangle_{\mathrm F}$ {\color{comment} \footnotesize \# Block gap}
\State $ \gamma \gets \min(1,\frac{g_i}{\frac{1}{N} \langle D, D-X^{(i)}P^{(i)}D\rangle_{\mathrm F}}) $ {\color{comment} \footnotesize \# Line-search}
%\State {\color{comment} \footnotesize \# W update}
\State $W_{t+1}$ $\gets$  $W_t + \gamma P^{(i)}D$ {\color{comment} \footnotesize \# W update}
%\State {\color{comment} \footnotesize \# Block update}
\State $Y^{(i)}_{t+1} \gets Y_t^{(i)} + \gamma D$ {\color{comment} \footnotesize \# Block update}
\EndFor
\end{algorithmic}
\end{algorithm}

\begin{figure*}[t]
  \begin{center}
    \mbox{}\vspace{-.6cm}\\
     \includegraphics[trim={5cm 5.5cm 3cm 6cm},clip,width=.95\textwidth]{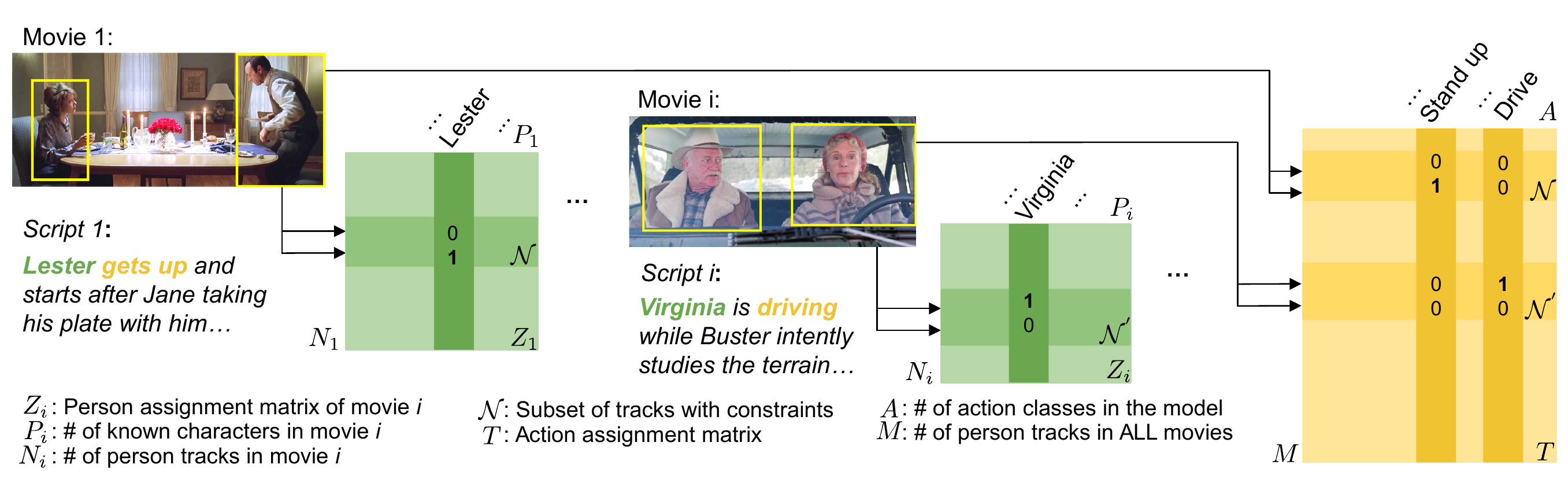}
\end{center}
\vspace{-.3cm}
\caption{Overview of the Person-Action weakly supervised model, see text for detailed explanations.}
\label{fig:method}
\end{figure*}

\section{Weakly supervised Person-Action model} 
\label{application}

We now describe an application of our large-scale discriminative clustering algorithm with weak-supervision. The goal is to
assign to each person track a name and an action. Both names and actions are mined from movie scripts.
For a given movie $i$, we assume to have $N_i$ automatically extracted person tracks as well as the parsing of a movie script into person names and action classes.
We also assume that scripts and movies have been roughly aligned in time.
In such a setup we can assign labels (\eg a name or an action) from a script section to a subset of tracks $\mathcal{N}$ 
from the corresponding time interval of a movie (see Figure~\ref{fig:method} for example).
In the following, we explain how to convert such form of weak supervision into a set of constraints on latent variables corresponding to the names and actions of people. 
We will also show how these constraints easily decompose into blocks.  
We denote $Z$ the latent variable assignment matrix for person names and $T$ for actions.

%
% which constraint on the latent variable we should define for training these weakly-supervised person tracks and how they decompose into small blocks. In
%the following, we note $Z$ the assignment matrix for person names and $T$ for actions.

%We will explain which constraint on the latent variable we should define for training these weakly-supervised person tracks and how they decompose into small blocks. In
%the following, we note $Z$ the assignment matrix for person names and $T$ for actions.

\subsection{Weak-supervision as constraints} \label{constraint}
We use linear constraints to incorporate weak supervision from movie scripts. In detail, we define constraints on subsets of person tracks that we call ``bags".
 %To do so we define subsets of tracks named bags and define constraints on them.
In the following we explain the procedure for construction of bags together with the definition of the appropriate constraints.
%different types of  types of bags and , that will be subject to supervision from the script. We explain here
%what are the different form of bags, how we construct them and what is their link to the linear constraints.
%Let us denote the set of tracks in a bag by $\mathcal{N}$.
%Our bags can take 3 different forms:
%if the bag is annotated by person $p$ then the bag is described by $(\mathcal{N},p, \emptyset)$,
%if it is annotated by action $a$ then we denote it as $(\mathcal{N}, \emptyset, a)$, 
%finally, if it is annotated by the person $p$ doing the action $a$ then we denote it as $(\mathcal{N}, p, a)$.
\vspace{-.3cm}

\paragraph{`At least one' constraint.} 
Suppose a script reveals the presence of a person $p$ in some time interval of the movie. We construct a set $\mathcal{N}$ with all person tracks in this interval.
As first proposed by \cite{bojanowski13finding}, we model that \emph{at least one} track in $\mathcal{N}$ is assigned to person $p$ by the following constraint
\begin{align}
   % \label{eq:mentionned-constraint}
   \label{eq:one-constraint}
    \sum_{n \in \mathcal{N}} Z_{np} \geq 1.
\end{align}
We can apply the same type of constraint when solving for action assignment $T$.
\vspace{-.3cm}

\paragraph{Person-Action constraint.} 
Scripts can also provide information that a person $p$ is performing an action $a$ in a scene. 
In such cases we can formulate stricter and more informative constraints as follows. 
We construct a set $\mathcal{N}$ containing all person tracks appearing in this scene. 
Following~\cite{bojanowski13finding}, we formulate a joint constraint on presence of a person performing a specific action as
\begin{align}
   % \label{eq:mentionned-constraint}
   \label{eq:person-action-constraint}
    \sum_{n \in \mathcal{N}} Z_{np}T_{na} \geq 1.
\end{align}
\vspace{-.3cm}

\paragraph{Mutual exclusion constraint.}
We also model that each person track can only be assigned to exactly one label. This restriction can be formalized by the mutual exclusion constraint
\begin{align}
  \label{eq:constraint-row}
Z1_{P} = 1_{N},
\end{align}
for $Z$ (\ie rows sum up to 1). Same constraint holds for $T$.
\vspace{-.3cm}

\paragraph{Background class constraint.} 
One of our contributions is a novel way of coping with the background class.
As opposed to previous work~\cite{bojanowski13finding}, our approach allows us to have background model that  does not require any external data.
Also it does not require a specific background class classifier as in~\cite{parkhi15it}. 

Our background class constraint can be seen as a way to supervise people and actions that are not mentioned in scripts.
We observe that tracks that are not subject to constraints from Eq.~\eqref{eq:one-constraint} and tracks that belong to crowded shots are likely to belong to the background class.
Let us denote by $\mathcal{B}$ the set of such tracks.
We impose that at least a certain fraction $\alpha \in [0,1]$ of tracks in $\mathcal{B}$ must belong to the background class.
Assuming that person label $p=1$ corresponds to the background, we obtain the following linear constraint (similar constraint can be defined for actions on $T$): 
\begin{align}
  \label{eq:background-constraint}
  \sum_{n \in \mathcal{B}} \ Z_{n1}  \geq \alpha \mid \mathcal{B} \mid .
\end{align}

\subsection{Person-Action model formulation}
\label{sec:jointmodel}
%We have so far described our cost function and the way of incorporating weak supervision by means of linear constraints. 
Here we summarize the complete formulation of the person and action recognition problems.  
%Using the constraints above, we first solve the character identification problem separately for each movie, and then use the recovered person identities
%as supervision for action recognition. % learned from all movies.
\vspace{-.3cm}

\paragraph{Solving for names.}
We formulate the person recognition problem as discriminative clustering, where $X_{1}$ are face descriptors: % extracted from person tracks: %are facial descriptors %descriptors: % (details given later):
%We formulate the person recognition problem as a standard quadratic problem, where $X_{1}$ are facial descriptors (details given later):
\begin{align}
    \label{eq:finalproblem}
    \min_{Z \in \{0,1\}^{N \times P}} & \quad \langle Z, A(X_{1},\lambda) Z\rangle_{\mathrm F}, \ \ \text{\footnotesize(Discriminative cost)}  \\
    \text{such that} &  \quad \begin{cases} 
                                \sum_{n \in \mathcal{N}} Z_{np} \geq 1, & \text{\footnotesize(At least one)}  \\
                                \sum_{n \in \mathcal{B}} Z_{n1} \geq \alpha \mid \mathcal{B} \mid, & \text{\footnotesize(Background)}   \\
                                Z1_{P} = 1_{N}. & \text{\footnotesize(Mutual exclusion)}
                              \end{cases}  \nonumber
\end{align}
\vspace{-.3cm}

\paragraph{Solving for actions.}
After solving the previous problem for names separately for each movie, we vertically concatenate all person name assignment matrices $Z$. We also define a single action assignment variable $T$ in $\{0, 1\}^{M \times A}$, where $M$ is the total number of tracks across all movies and 
$X_{2}$ are action descriptors (details given later). We formulate our action recognition problem as a large QP:
\begin{align}
  \label{eq:finalproblemaction}
  \min_{T \in \{0,1\}^{M \times A}} & \quad \langle T, A(X_{2},\mu) T\rangle_{\mathrm F}, \ \ \text{\footnotesize(Discriminative cost)}   \\ 
  \text{such that} & \quad  \begin{cases} 
                              \sum_{n \in \mathcal{N}} \ T_{na} \geq 1, & \text{\footnotesize(At least one)} \\
                              \sum_{n \in \mathcal{N}} Z_{np}T_{na}  \geq 1, & \text{\footnotesize(Person-Action)} \\
                              \sum_{n \in \mathcal{B}} \ T_{n1}  \geq \beta \mid \mathcal{B} \mid, & \text{\footnotesize(Background)} \\
                              T 1_{A} = 1_{M}. & \text{\footnotesize(Mutual exclusion)}
                            \end{cases}  \nonumber 
\end{align}

\paragraph{Block-Separable constraints.} The set of linear constraints on the action assignment matrix T is block separable since
each movie has it own set of constraints, i.e.\ there are no constraints spanning multiple movies. Therefore, we can fully demonstrate here the power of our large-scale discriminative clustering optimization (Algorithm~\ref{CHalgorithm}). % we define each individual movie as one block.

%\paragraph{Block-Separable constraints.} We can then smoothly apply our algorithm~\ref{CHalgorithm} to solve this problem on a large-scale dataset of movies. Indeed, the set of linear constraints on T are block separable since
%each movies has it own defined constraint: there is no constraint overlapping between movies. So, in our case, each block would be a movie block.

%\input{optimization}

%%%%%%%%%%%%%
%%% SETUP %%%
%%%%%%%%%%%%%

\section{Experimental Setup}

\subsection{Dataset}
Our dataset is composed of 66 Hollywood feature-length movies (see the list in Appendix) that we obtained from either BluRay or DVD.
For all movies, we downloaded their scripts (on~\url{www.dailyscript.com})
that we temporally aligned with the videos and movie subtitles using the method described in \cite{laptev08learning}. The total number of frames in all 66 movies is 11,320,252. The number of body tracks detected across all movies (see \ref{bodytracks} for more details) is $M = 201874$.

\subsection{Text pre-processing} \label{nlp}
To provide weak supervision for our method we process movie scripts to extract occurrences of the 13 most frequent action classes:
%of the following actions in scripts:  We need to extract from the script descriptions, when one of these following actions: 
\texttt{Stand Up}, \texttt{Eat}, \texttt{Sit Down}, \texttt{Sit Up}, \texttt{Hand Shake}, \texttt{Fight}, \texttt{Get Out of Car}, \texttt{Kiss}, \texttt{Hug}, \texttt{Answer Phone}, \texttt{Run}, \texttt{Open Door} and \texttt{Drive}. % happen.
To do so, we collect a corpus of movie scripts different from the set of our 66 movies and train simple text-based action classifiers using linear SVM and a TF-IDF representation of words composed of uni-grams and bi-grams. 
%a linear SVM on this corpus using a TF-IDF representation of words composed of uni-grams and bi-grams.
After retrieving actions in our target movie scripts, we also need to identify who is performing the action. We used spaCy~\cite{honnibalspacy} to parse every sentence classified as describing one of the 13 actions
and get every subject for each action verb. 
%We could potentially increase the action vocabulary size, but we chose to restrict to these as our movie scripts dataset does not contain enough occurrence for other actions.

%%sentence classified as describing one of the 13 actions. After that, we looked for any main character name his associated verb using the syntactic tree. Then,
%%we manually construct a list of lemmatized verbs with similar semantic (for example, the list would be: \textit{'fight','kick','punch'} for action \texttt{Fight}) 
%%for all the actions. Finally we need to check if the associated lemmatized verb belongs or not to the action classified by the SVM. For example,
%%if the sentence: '\textit{Jack is exiting the room, then Lana rushes toward him.}' is classified as a \texttt{Run} sentence. Only 'Lana' will be output as 
%%'Jack' is linked to the verb \textit{exit} which is not in the list associated to \texttt{Run}. 'Lana' is linked to the verb \textit{rush} which is in the list of verbs for the class \texttt{Run}.

\subsection{Person detection and Features}
\paragraph{Face tracks.}  To obtain tracks of faces in the video, we run the multi-view face detector~\cite{mathiasdpm} based on the DPM model~\cite{dpm}.
We then extract face tracks using the same method as in~\cite{everingham06hello,sivic09who}. 
 For each detected face, we compute facial landmarks~\cite{sivic09who} followed by the face alignment and resizing of face images to 224x224 pixel. We use pre-trained vgg-face features~\cite{parkhideepface} to 
extract descriptors for each face. We kept the features of dimension 4096 computed by the network at the last fully-connected
layer that we $L_{2}$ normalized. For each face track, we choose the top K (in practice, we choose K=5) faces that have the best facial landmark confidence. Then we represent each track by averaging the features of the top K faces.% retrieved.

\paragraph{Body tracks.} \label{bodytracks} 
To get the person body tracks, we run the Faster-RCNN network with VGG-16 architecture fine-tuned on VOC 07 \cite{shaoqingfasterrcnn}. Then we track bounding boxes using the same tracker as used to obtain face tracks.
To get person identity for body tracks, we greedily link each body track to one face track by maximizing a spatio-temporal bounding box overlap measure.
However if a body track does not have an associated face track as the actor's face may look away from the camera, we cannot obtain its identity.
Such tracks can be originating from any actor in the movie. %Thus we possibly consider the body track to be from anyone.
To capture motion features of each body track, we compute bag-of-visual-words representation of dense trajectory descriptors~\cite{wang13action} inside the bounding boxes defined by the body track. We use 4000 cluster centers for each of the HOF, MBHx and MBHy channels.
%The appearance naturally helps when it comes to classifying some actions. 
In order to capture appearance of each body track we extract ResNet-50 \cite{he16resnet} pre-trained on ImageNet. For each body bounding box, we compute the average RoI-pooled \cite{shaoqingfasterrcnn} feature map of the last convolutional
layer within the bounding box, which yields a feature vector of dimension 2048 for each box.
We extract a feature vector every 10th frame, average extracted feature vectors over the duration of the track and $L_{2}$ normalize.
Finally, we concatenate the dense trajectory descriptor and the appearance descriptor resulting in a 14028-dimensional descriptor for each body track. 
%The final vector of size 14028 representing the body track is a concatenation of the dense trajectories and appearance features.

%%%%%%%%%%%%%%%%%%%
%%% EXPERIMENTS %%%
%%%%%%%%%%%%%%%%%%%

\section{Evaluation}

\subsection{Evaluation of person recognition}

\begin{table}[t]
  \setlength{\tabcolsep}{3pt}
    \centering
    \vspace{0pt}
    \resizebox{\columnwidth}{!}{
    \begin{tabular}{@{}lccc@{}}
      \toprule
      Method                                              & Acc. & Multi-Class AP & Background AP\\
      \midrule
      Cour \etal \cite{cour09learning}            & $48 \%$ & $63 \%$ & $-$ \\
      Sivic \etal \cite{sivic09who}               & $49 \%$ & $63 \%$ & $-$ \\
      Bojanowski \etal \cite{bojanowski13finding} & $57 \%$ & $75 \%$ & $51 \%$\\
      Parkhi \etal \cite{parkhi15it}              & $74 \%$ & $93 \%$ & $75 \%$\\
      \textbf{Our method}                                 & $\mathbf{83 \%}$ & $\mathbf{94 \%}$ & $\mathbf{82} \%$ \\
      \bottomrule
    \end{tabular}
     }
    \caption{Comparison on the Casablanca benchmark \cite{bojanowski13finding}.}
      \label{table:casablanca-comparison}
  \vspace{0.2cm}      
     \begin{tabular}{@{}l ccccc@{}}
    \toprule
    Episode & 1 & 2 & 3 & 4 & 5 \\
    % METHOD & \# Training movie & St.U. & E. & S.D.  & Si.U. & H.S. & F. & G.C. & K. & H. & A. & R. & O.D. & D. & mAP \\
    \midrule
    Sivic \etal \cite{sivic09who}   & 90 & 83 & 70 & 86 & 85 \\
    Parkhi \etal \cite{parkhi15it} & \textbf{99} & 90 & 94 & 96 & \textbf{97} \\
    \textbf{Ours} & 98 & \textbf{98} & \textbf{98} & \textbf{97} & \textbf{97} \\
    \bottomrule
  \end{tabular}
 
  \vspace{0.1cm}
  \caption{Comparison on the Buffy benchmark \cite{sivic09who} using AP.}
  \vspace{0.2cm}
  \label{table:buffy-comparison}
      \begin{tabular}{@{}l cccccccc@{}}
    \toprule
    $\alpha$ & 0 & 0.1 & 0.2 & 0.3 & 0.4 & 0.5 & 0.75 & 1.0 \\
    \midrule
    Accuracy & 58 & 58 & 70 & 82 & \textbf{84} & 83 & 76 & 55 \\
    AP & 86 & 87 & 90 & \textbf{94} & \textbf{94} & 93 & 85 & 58 \\
    \bottomrule
  \end{tabular}

  \vspace{0.2cm}
  \caption{Sensitivity to hyper-parameter $\alpha$~\eqref{eq:background-constraint} on Casablanca.}
  \label{table:alpha-experiment}
  \vspace{-0.7cm}
\end{table}

\begin{figure*}[t]
  \centering
  \includegraphics[width=\linewidth]{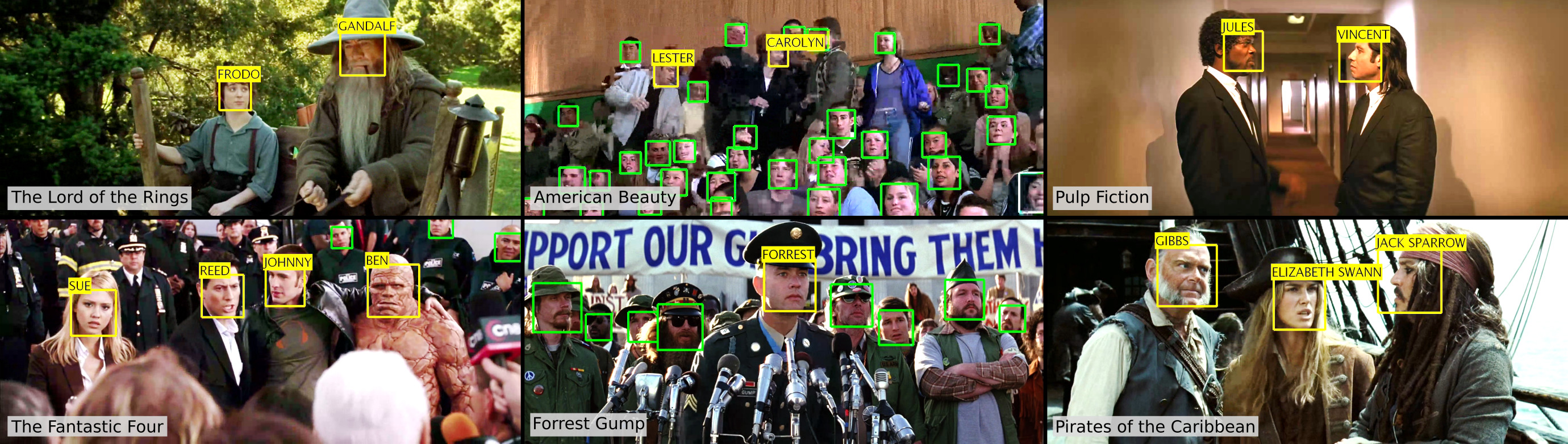}
  \caption{
    Qualitative results for face recognition. 
    Green bounding boxes are face tracks correctly classified as background characters.
  }
  \label{fig:qualitative-faces}
  \vspace{-.2cm}
\end{figure*}

We compare our person recognition method to several other methods on the Casablanca benchmark from~\cite{bojanowski13finding} and
the Buffy benchmark from~\cite{sivic09who}.
All methods are evaluated on the same inputs: same face tracks, scripts and characters. Table~\ref{table:casablanca-comparison} shows the Accuracy (Acc.) and Average Precision (AP) of
our approach compared to other methods on the Casablanca benchmark~\cite{bojanowski13finding}. In particular we compare to Parkhi \etal~\cite{parkhi15it} which is a strong baseline using the same CNN face descriptors as in our method.
%features is a strong recent work, which reported previous state-of-the-art results on this benchmark using same CNN features.
We also show the AP of classifying the background character class (Background AP). % in Casablanca.
We compare in Table~\ref{table:buffy-comparison} our approach to other methods~\cite{parkhi15it,sivic09who} reporting results on season 5 of the TV series ``Buffy the Vampire Slayer".
Both of these methods~\cite{parkhi15it,sivic09who} use speaker detection to mine additional strong (but possibly incorrect) labels from the script, which we also incorporate (as additional bags) to make the comparison fair.
Our method demonstrates significant improvement over the previous results. It also outperforms other methods on the task of classifying background characters.
Finally, Table~\ref{table:alpha-experiment} shows the sensitivity to hyper-parameter $\alpha$ from the background constraint~\eqref{eq:background-constraint} on the Casablanca benchmark.
Note that in contrast to other methods, our background model does not require supervision for the background class. 
This clearly demonstrates the advantage of our proposed background model. For all experiments the hyper-parameter $\alpha$ of the background constraint~\eqref{eq:background-constraint} was set to $30 \%$.
Figure~\ref{fig:qualitative-faces} illustrates our qualitative results for character recognition in different movies.

\subsection{Evaluation of action recognition}

\begin{table*}[h]
  \begin{center}
    % \scriptsize{
    %   \scalebox{1}{
    %     \begin{tabular}{@{}lcccccccccccccccc@{}}
    %       \hline
    %       ACTION & \# Movies  & Other & St.U. & E. & S.D.  & Si.U. & H.S. & F. & G.C. & K. & H. & A. & R. & O.D. & D. & Total \\
    %       \hline
    %       \small Groundtruth & 5 & 14532 & 146 & 24 & 112 & 19 & 28 & 90 & 26 & 47 & 74 & 28 & 277 & 131 & 59 & 15593\\
    %       \small Constraint & 66 & $\emptyset $ & 237 & 85 & 146 & 46 & 49 & 70 & 81 & 244 & 44 & 99 & 156 & 208 & 169 & 1634 \\
    %       \hline
    %     \end{tabular}
    %   }
    % }
    % \caption{Action recognition groundtruth and constraint statistics}
    % \label{table:stats}
  
  \resizebox{\textwidth}{!}{
  
  \begin{tabular}{@{}lcc|cccccccccccccccc @{}}
    \toprule
    METHOD & \# movies & Joint-Model & St.U. & E. & S.D.  & Si.U. & H.S. & F. & G.C. & K. & H. & A. & R. & O.D. & D. & mAP \\
    % METHOD & \# Training movie & St.U. & E. & S.D.  & Si.U. & H.S. & F. & G.C. & K. & H. & A. & R. & O.D. & D. & mAP \\
    \midrule
    \small (a) Random & $\emptyset $  & No  & 0.9 & 0.1 & 0.7 & 0.1 & 0.1 & 0.6 & 0.2 & 0.3 & 0.5 & 0.2 & 1.8 & 0.8 & 0.4 & 0.5 \\
    \small (b) Script only & $\emptyset $ & No & 3.0 &   4.3  &  5.5  &  2.8  &  4.7  &  2.5  &  1.6  &  11.3  &  4.2 &   1.4  &  13.7 &   3.1  &  3.0 &  4.7\\
    \small (c) Fully-supervised  &  4  & No &  21.2 &  0.2  & 22.2   & 0.9  &  0.6  &  7.3  &  1.4  &  1.9  &  \textbf{4.5}  & 2.0 &  33.2  &  18.5  &  6.3 & 9.3 \\
    \small (d) Few training movies &  5 & Yes  &  22.6  & 9.6   &  15.6  &  \textbf{8.1}  &  9.7   & 6.1  &  1.0 &   6.0  &  2.1  &  4.2   & 44.0  &  16.2  &  15.9 & 12.4 \\ 
    \small (e) No Joint Model  & 66   & No   & 10.7  &  7.0  &  17.1  &  7.3  &  \textbf{18.0} &   \textbf{12.6}  & \textbf{2.0}   & \textbf{14.9}  &  3.6  &  \textbf{5.8} &   24.4  &  14.2 &   24.9  & 12.5  \\
    % \small (f) Full setup ($T$) & 30 & 19.3 & 10.3  & 22.8 & 11.8 &  15.0 &  6.0 & 1.1 & 10.6 & 2.6 &  6.2  &  48.4 &  17.0  &  28.9 & 15.4\\
    % \small (f) Full setup ($T$) & 40 & 21.6 & 9.8  & 24.3 & 13.6 &  11.8 &  7.0 & 1.2 & 10.6 & 2.5 &  6.8  &  48.0 &  16.5  &  30.8 & 15.7\\
    %  \small (f) Full setup ($T$) & 50 & 21.6 & 9.8  & 24.3 & 13.6 &  11.8 &  7.0 & 1.2 & 10.6 & 2.5 &  6.8  &  48.0 &  16.5  &  30.8 & 16.3\\
     \small (f) Full setup &  66  & Yes   &  \textbf{27.0} &  \textbf{9.8}  &  \textbf{28.2}   & 6.7  &  7.8  &  5.9  &  1.0  &  12.9  &  1.7  & 5.7 &   \textbf{56.3}  &  \textbf{21.3}  &  \textbf{29.7} & \textbf{16.4} \\
    \bottomrule
  \end{tabular}
  }
  \end{center}
    \vspace{-0.1cm}
  \caption{Average Precision of actions evaluated on 5 movies. (St.U: \texttt{Stand Up}, E.: \texttt{Eat}, S.D: \texttt{Sit Down}, Si.U.: \texttt{Sit Up},
  H.S: \texttt{Hand Shake}, F.: \texttt{Fight}, G.C.: \texttt{Get out of Car}, K.: \texttt{Kiss}, H.: \texttt{Hug}, A.: \texttt{Answer Phone}, R.: \texttt{Run}, O.D.: \texttt{Open Door}, D.: \texttt{Drive})}
  \label{table:ac-comparison}

\end{table*}

\begin{figure}

  \centering
     \includegraphics[width=0.75\linewidth]{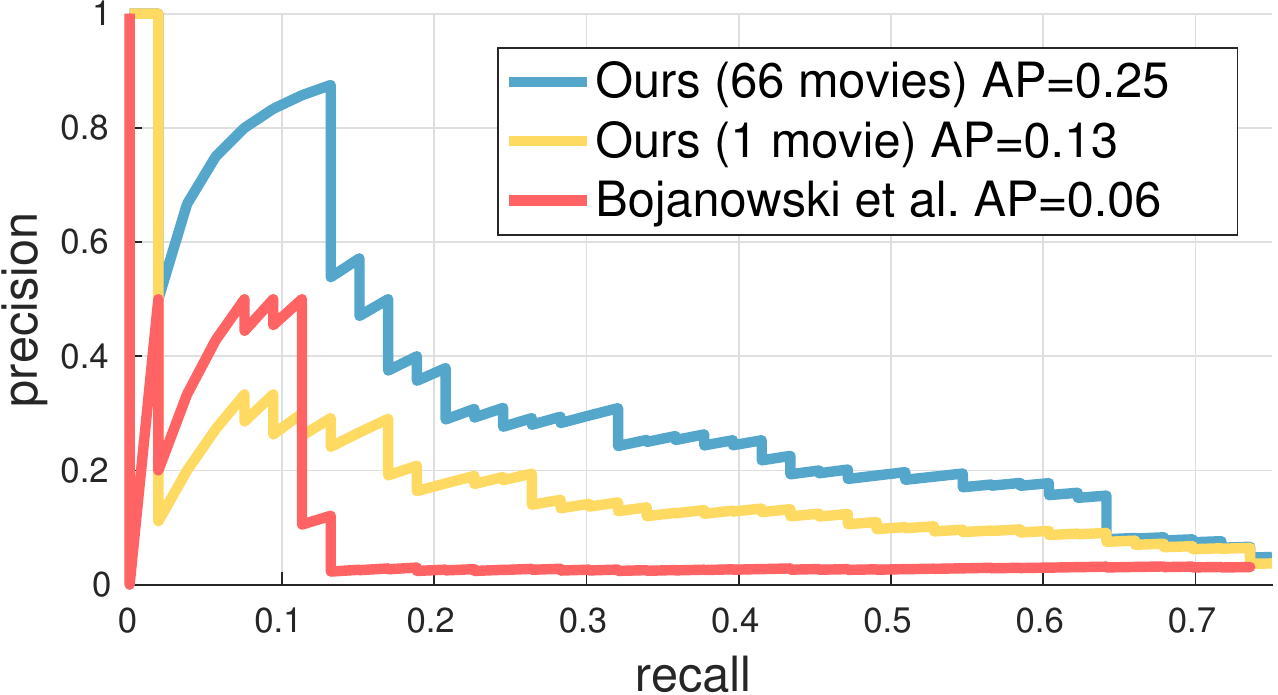}
  \caption{\small PR curves of action SitDown from Casablanca.}
  \label{fig:pr-sitdown}

  \centering
     \includegraphics[width=0.75\linewidth]{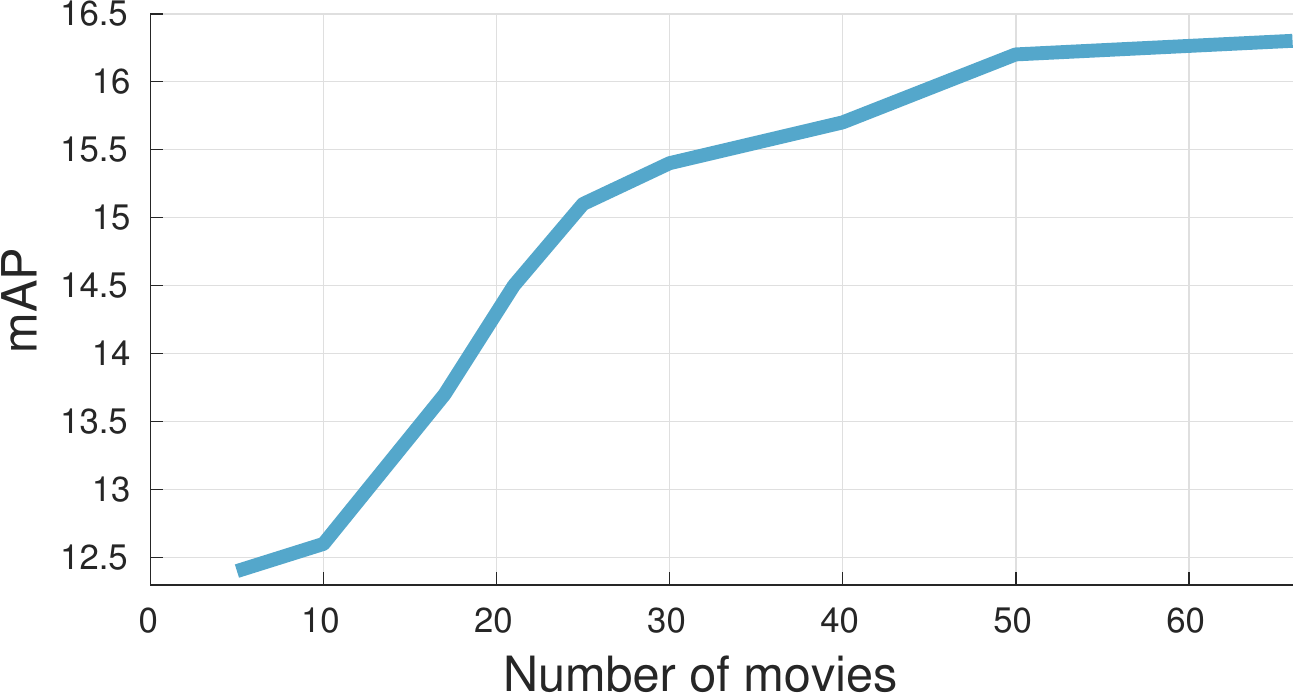}
  \caption{\small Action recognition mAP with increasing number of training movies.}
  \label{fig:map-curve}
\vspace{-.5cm}
\end{figure}

First, we compare our method to Bojanowski \etal 2013~\cite{bojanowski13finding}. Their evaluation uses different body tracks than ours, we design here
an algorithm-independent evaluation setup.
%Since the method~\cite{bojanowski13finding} was evaluated on two movies with two actions each,
We compare our model using the Casablanca movie and the \texttt{Sit Down} action.
For the purpose of evaluation, we have manually annotated all person tracks in the movie and then manually labeled whether or not they contain the \texttt{Sit Down} action.
%with manually annotated head tracks on the only action \texttt{Sit Down} we have in common.
Given this ground truth, we assess the two models in a similar way as typically done in object detection.
%Hence, the our experimental setup is independent of body tracks.
Figure \ref{fig:pr-sitdown} shows a
precision-recall curve evaluating recognition of the \texttt{Sit Down} action. We show our method trained on Casablanca only (as done in~\cite{bojanowski13finding})
and then on all 66 movies. Our method trained on Casablanca is already better than~\cite{bojanowski13finding}. The improvement becomes even more evident when training our method on all 66 movies.

To evaluate our method on all 13 action classes, we use five movies (American Beauty, Casablanca, Double Indemnity, Forrest Gump and Fight Club).
For each of these movies we have manually annotated all person tracks produced by our tracker according to 13 target action classes and the background action class.
We assume that each track corresponds to at most one target action.
In rare cases where this assumption is violated, we annotate the track by one of the correct action classes. 

\begin{table}[t]
  \setlength{\tabcolsep}{3pt}
    \centering
    \vspace{0pt}
      \resizebox{0.4\textwidth}{!}{
    \begin{tabular}{@{}lcccc@{}}
      \toprule
      Method                                              & R@1 & R@5 & R@10 & Median Rank \\
      \midrule
      Yu \etal \cite{yu16videocaptioning}            & $3.6\%$ & $14.7 \%$ & $23.9 \%$ & $\mathbf{50}$ \\
      Levi \etal \cite{levi16videodescription}               & $4.7 \%$ & $15.9 \%$ & $23.4 \%$ & $64$ \\
      \textbf{Our baseline}                                 & $\mathbf{7.3 \%}$ & $\mathbf{19.2 \%}$ & $\mathbf{27.1} \%$  & $52$\\
      \bottomrule
    \end{tabular}
    }

    \caption{Baseline comparison against winners of the LSMDC2016 movie clip retrieval challenge}
      \label{table:baseline-comparison}
      \vspace{-.5cm}
\end{table}

\begin{figure*}[t]
  \centering
  \includegraphics[width=\linewidth]{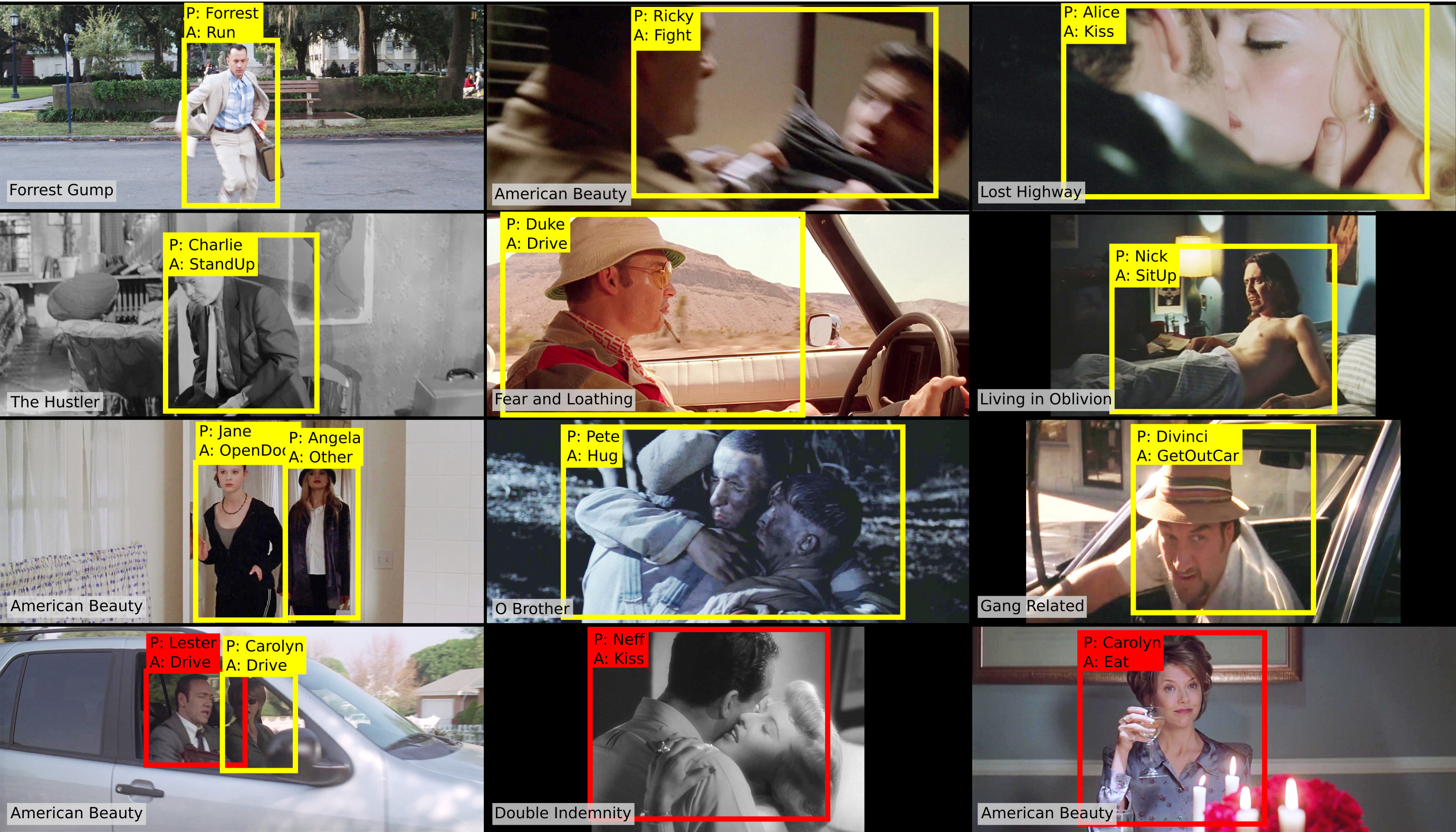}
  \caption{
    Qualitative results for action recognition. P stands for for the name of the character and A for the action performed by P.
    Last row (in red) shows mislabeled tracks with high confidence 
    (\eg hugging labeled as kissing, sitting in a car labeled as driving).
  }
  \label{fig:qualitative-action}
  \vspace{-.3cm}
\end{figure*}

In Table~\ref{table:ac-comparison} we compare results of our model to different baselines. % using the same on the same five movies.
The first baseline (\textbf{a}) corresponds to the random assignment of action classes.
The second baseline (\textbf{b}) Script only uses information extracted from the scripts: each time an action appears in a bag, all person tracks in this bag are then simply annotated with this action. 
Baseline (\textbf{c}) is using our action descriptors but trained in a fully supervised set-up on a small subset of annotated movies.
To demonstrate the strength of this baseline we have used the same action descriptors on the LSMDC2016\footnote{https://sites.google.com/site/describingmovies/lsmdc-2016}
movie clip retrieval challenge. This is the largest public benchmark~\cite{rohrbach15dataset} related to our work that considers movie data (but without person localization as we do in our work).
Table~\ref{table:baseline-comparison} shows our features employed in simple CCA method as done in~\cite{levi16videodescription} achieving state-of-the-art on this benchmark.
The fourth baseline (\textbf{d}) is our method train only \emph{using the five evaluated movies}. 
The fifth baseline (\textbf{e}) is our model without the joint person-action constraint (\ref{eq:person-action-constraint}), but still trained on all 66 movies. 
Finally, the last result (\textbf{f}) is from our model using all the 66 training movies and person-action constraints (\ref{eq:person-action-constraint}).
Results demonstrate that optimizing our model on more movies brings the most significant improvement to the final results.
We confirm the idea from~\cite{bojanowski13finding} that adding the information of who is performing the action in general helps identifying actions.
However we also notice it is not always true for actions with interacting people such as: \texttt{Fight}, \texttt{Hand Shake}, \texttt{Hug} or \texttt{Kiss}.
Knowing who is doing the action does not seems to help for these actions.
Figure~\ref{fig:map-curve} shows improvements in action recognition when gradually increasing the number of training movies. Figure~\ref{fig:qualitative-action} shows qualitative results of our model on different movies. The statistics about the ground truth and constraints together with additional results are provided in {\bf Appendix}.

\vspace{-.5cm}
\vspace{-0.8cm}
\section{Conclusion}
\vspace{-.3cm}
We have proposed an efficient online optimization method based on the Block-Coordinate Frank-Wolfe algorithm.
We use this new algorithm to scale-up discriminative clustering model in the context of weakly-supervised person and action recognition in feature-length movies.
Moreover, we have proposed a novel way of handling the background class, which does not require collecting background class data as required by the previous approaches, and leads to better performance for person recognition.
%By allowing model to scale, we jointly trained it on a large number of feature-length movies. 
In summary, the proposed model significantly improves action recognition results on 66 feature-length movies.
The significance of the technical contribution goes beyond the problem of person-action recognition as the proposed optimization algorithm can scale-up other problems recently tackled by discriminative clustering.
Examples include: unsupervised learning from narrated instruction videos~\cite{alayrac16unsupervised}, text-to-video alignment~\cite{bojanowski15weakly}, co-segmentation~\cite{joulin10discriminative}, co-localization in videos and images~\cite{tang14efficient} or instance-level segmentation~\cite{seguin16instance}, which can be now scaled-up to an order of magnitude larger datasets.
\vspace{-.3cm}

{\footnotesize \paragraph{Acknowledgments.} This work has been supported by ERC grants ACTIVIA (no.\
307574) and LEAP (no.\ 336845), CIFAR Learning in Machines $\&$ Brains
program, ESIF, OP Research, development and education Project IMPACT
No.\ CZ$.02.1.01/0.0/0.0/15\_003/0000468$ and a Google Research Award.}

{\small
\bibliographystyle{ieee}
\bibliography{master-biblio}
}
\clearpage

\appendix

\section{Appendix}
This appendix contains details and supplementary results to the main paper.

\subsection{Slack variables}
\label{sec:slack}
To account for imprecise information in movie scripts, we add slack variables to our constraints.
We penalyze the values of slack variables with the $L_2$ penalty. The slack-augmented constraints are defined as:
%The constraints we use in our experiments are defined as: 
\begin{align}
    \sum_{n \in \mathcal{N}} \ Z_{np}  \geq 1 - \xi, \label{eq:slackZ}\\
    \sum_{n \in \mathcal{N}} \ T_{na}  \geq 1 - \xi, \label{eq:slackT}\\
    \sum_{n \in \mathcal{N}} \ Z_{np} \ T_{na}  \geq 1  - \xi \label{eq:slackZT},
\end{align}
where $\xi$ is the slack variable.

\subsection{Lower bound}
\label{sec:lowerbound}
In practice, we noticed that modifying the value of the lower bound in constraints~(\ref{eq:slackZ}),~(\ref{eq:slackT}),~(\ref{eq:slackZT}) from 1 to a higher value can significantly improve the performance of the algorithm. 
The constraints we use become:
\begin{align}
   \sum_{n \in \mathcal{N}} \ Z_{np}  \geq \alpha_{1} - \xi , \\
   \sum_{n \in \mathcal{N}} \ T_{na}  \geq \alpha_{2} - \xi, \\
   \sum_{n \in \mathcal{N}} \ Z_{np} \ T_{na}  \geq \alpha_{2} - \xi,   
\end{align}
where $\alpha_{1}, \alpha_{2} \in \mathbb{R_+}$ are hyper-parameters. \\

\subsection{Combining face and body tracks}
\label{sec:tracks}
% We describe here in detail how we linked body tracks to face tracks.
 
Let's denote $a_{1}, a_{2}, ..., a_{n}$, $n$ faces tracks in the current shot and $b_{1}, b_{2}, ..., b_{m}$ the $m$ body tracks in this same shot (we assume $ m \geq n $).
We want to model that each face track is associated to at most one body track but a body track does not necessary
have a face track, as the face of a person may not always be visible.
Let's also define the following overlap measure $O$ between a face track $a$ and a body track $b$.
If $\mathcal{A}$ is a set of all frames of the track $a$ and $a(t)$, $b(t)$ are bounding boxes of tracks $a$ and $b$ at frame $t$, we have: 
 \begin{align}
O(a,b) = \sum_{t \in \mathcal{A}} \ \frac{Area(a(t) \cap b(t))}{Area(a(t))}.
 \end{align}
 We compute the overlap for all possible pairs $O(a_{i},b_{j})$, where $i \in [1,n]$ and $j \in [1,m]$.
 Then we associate each face track $a_{i}$ with the body track $b_{j}$ that maximizes $O(a_{i},b_{j})$. Finally, for each body track $b_{j}$ we 
 either do not have any associated face track (then the body track won't have a match) or have multiple face tracks $a_{i}$ associated to it. In the latter case, we match
 the body track $b_{j}$ with the face track $a_{i}$ that maximizes $O(a_{i},b_{j})$.

\begin{table}[t]
  \scalebox{0.9}{
      \begin{tabular}{@{}l ccccccccc@{}}
    \toprule
    $\beta$ & 0 & 0.1 & 0.2 & 0.3 & 0.4 & 0.5 & 0.6 & 0.75 & 0.8 \\
    \midrule
    mAP & 15.0 & 15.7 & 15.9 & 15.8 & 16.6 & 16.1 & 16.2 & 16.0 & 15.5 \\
    \bottomrule
  \end{tabular}
  }
  \caption{Influence of the hyper-parameter $\beta$~\eqref{eq:background-constraint} for action recognition.}
     \label{table:beta-experiment}
\end{table}

\subsection{Sensitivity to the background constraint hyperparameter for action recognition}
\label{sec:beta-exp}
% We describe here in detail how we linked body tracks to face tracks.
Table~\ref{table:beta-experiment} shows the low sensitivity of the action recognition results to the $\beta$~\eqref{eq:finalproblemaction} hyper-parameter on the action recognition results.

\subsection{Additional dataset statistics}
\label{sec:stats}
 Table~\ref{table:stats} provides the number of action constraints we extracted from 66 movie scripts. It also shows the number of ground truth intervals for each action we obtained by an exhaustive manual annotation of human actions in five testing movies.

Our dataset contains the following 66 movies: 
%\section{Movie list}
%\label{sec:movielist} 
\normalsize{American Beauty, As Good As It Gets, Being John Malkovich, Big Fish, Bringing Out the Dead, Bruce the Almighty, Casablanca, Charade, Chasing Amy, Clerks, Crash, Dead Poets Society, Double Indemnity, Erin Brockovich, Fantastic Four, Fargo, Fear and Loathing in Las Vegas, Fight Club, Five Easy Pieces, Forrest Gump, Gandhi, Gang Related, Get Shorty, Hudsucker Proxy, I Am Sam, Independence Day, Indiana Jones and the Last Crusade, It Happened One Night, Jackie Brown, Jay and Silent Bob Strike Back, LA Confidential, Legally Blonde, Light Sleeper, Little Miss Sunshine, Living in Oblivion, Lone Star, Lost Highway, Men In Black, Midnight Run, Misery, Mission to Mars, Moonstruck, Mumford, Ninotchka, O Brother, Pirates of the Caribbean Dead Mans Chest, Psycho, Pulp Fiction, Quills, Raising Arizona, Rear Window, Reservoir Dogs, The Big Lebowski, The Butterfly Effect, The Cider House Rules, The Crying Game, The Godfather, The Graduate, The Grapes of Wrath, The Hustler, The Lord of the Rings The Fellowship of the Ring, The Lost Weekend, The Night of the Hunter, The Pianist, The Princess Bride, Truman Capote.}

\begin{table*}[t]
  \begin{center}
     \scriptsize{
      \scalebox{1.15}{
         \begin{tabular}{@{}lcccccccccccccccc@{}}
             \toprule
           ACTION & \# movies  & Other & St.U. & E. & S.D.  & Si.U. & H.S. & F. & G.C. & K. & H. & A. & R. & O.D. & D. & Total \\
    \midrule
          \small Ground truth & 5 & 14532 & 146 & 24 & 112 & 19 & 28 & 90 & 26 & 47 & 74 & 28 & 277 & 131 & 59 & 15593\\
          \small Constraints & 66 & $\emptyset $ & 237 & 85 & 146 & 46 & 49 & 70 & 81 & 244 & 44 & 99 & 156 & 208 & 169 & 1634 \\
   \bottomrule
         \end{tabular}
      }
     }
     \caption{Action recognition ground truth and constraint statistics. (St.U: \texttt{Stand Up}, E.: \texttt{Eat}, S.D: \texttt{Sit Down}, Si.U.: \texttt{Sit Up},
  H.S: \texttt{Hand Shake}, F.: \texttt{Fight}, G.C.: \texttt{Get out of Car}, K.: \texttt{Kiss}, H.: \texttt{Hug}, A.: \texttt{Answer Phone}, R.: \texttt{Run}, O.D.: \texttt{Open Door}, D.: \texttt{Drive})}
     \label{table:stats}
 \end{center}
\end{table*}

\end{document}